\pdfoutput=1

\documentclass[11pt]{article}

\usepackage[preprint]{acl}
 \usepackage{amssymb}
\usepackage{times}
\usepackage{latexsym}
\usepackage{amsmath}
\usepackage[T1]{fontenc}
\usepackage{tocloft}
\usepackage[utf8]{inputenc}

\usepackage{microtype}
\usepackage{booktabs}
\usepackage{inconsolata}

\usepackage{graphicx}

\usepackage{subcaption}
\usepackage{caption}
\usepackage[utf8]{inputenc}
\usepackage{array}
\usepackage{tabularx}
\usepackage{xcolor}
\usepackage{booktabs}
\usepackage[font=small,skip=5pt]{caption}
\usepackage{enumitem}
\usepackage{makecell}
\usepackage{multirow}
\usepackage{xcolor}  
\usepackage{titlesec}
\usepackage[most]{tcolorbox} 
\usepackage{xcolor}
\usepackage{lipsum} 
\usepackage{algorithm}
\usepackage{algpseudocode} 
\definecolor{titlebg}{RGB}{60,60,60}   
\definecolor{boxbg}{RGB}{245,245,245}  

\usepackage{amsmath}
\usepackage{enumitem}
\setlist[itemize]{itemsep=1pt, topsep=2pt, parsep=0pt, partopsep=0pt}

\AtBeginDocument{%
  \everydisplay\expandafter{\the\everydisplay\small}%

  \setlength{\abovedisplayskip}{4pt plus 2pt minus 2pt}%
  \setlength{\belowdisplayskip}{4pt plus 2pt minus 2pt}%
  \setlength{\abovedisplayshortskip}{3pt plus 2pt minus 2pt}%
  \setlength{\belowdisplayshortskip}{3pt plus 2pt minus 2pt}%
}

\title{Beyond Experience Retrieval: Learning to Generate Utility-Optimized Structured Experience for Frozen LLMs}
\author{
  \textbf{Xuancheng Li\textsuperscript{1,2,$\dagger$}},
  \textbf{Haitao Li\textsuperscript{1,2,$\dagger$}},
  \textbf{Yujia Zhou\textsuperscript{1,2}},
  \textbf{Yiqun Liu\textsuperscript{1}},
  \textbf{Qingyao Ai\textsuperscript{2,1,*}} \\
  \textsuperscript{1}Department of Computer Science and Technology, Tsinghua University, Beijing, China \\
  \textsuperscript{2}Quancheng Laboratory \\
  \texttt{lixuancheng23@mails.tsinghua.edu.cn} \\
  \texttt{liht22@mails.tsinghua.edu.cn, zhouyujia@mail.tsinghua.edu.cn} \\
  \texttt{yiqunliu@tsinghua.edu.cn, aiqy@tsinghua.edu.cn}
}
\begin{document}
\maketitle
\begingroup
\renewcommand\thefootnote{\fnsymbol{footnote}}
\footnotetext[2]{Equal contribution.}
\footnotetext[1]{Corresponding author.}
\endgroup
\begin{abstract}
Large language models (LLMs) are largely static and often redo reasoning or repeat mistakes. Prior experience reuse typically relies on external retrieval, which is similarity-based, can introduce noise, and adds latency. We introduce \textbf{SEAM} (\textbf{S}tructured \textbf{E}xperience \textbf{A}dapter \textbf{M}odule), a lightweight, executor-specific plug-in that stores experience in its parameters and generates a structured, instance-tailored experience entry in a single forward pass to guide a frozen LLM executor. SEAM is trained for utility via executor rollouts and GRPO while keeping the executor frozen, and can be further improved with logged-success SFT after deployment. Experiments on mathematical reasoning benchmarks show consistent accuracy gains across executors with low overhead. Extensive ablation and analysis further elucidate the mechanisms underlying SEAM’s effectiveness and robustness.\footnote{We release our code at \url{https://github.com/Cocobalt/SEAM}.}

\end{abstract}

\section{Introduction}

Large language models (LLMs) have achieved strong performance across diverse tasks, including machine translation, information retrieval, and mathematical reasoning \citep{ahn2024large,naveed2025comprehensive,shi2025deepresearchsystematicsurvey}. 
Yet, they remain largely \emph{static}: when confronted with new problems, LLMs often reason from scratch, re-explore familiar solution paths, and repeat avoidable mistakes \citep{feng2025get,fan2025if,yu2025stateful}. 
In contrast, humans distill procedural experience from past problem solving and apply it to new situations, improving efficiency and avoiding prior failures \citep{efklides2009role,kolb2014experiential,usher1993experiential}. 
Ideally, an agent should also accumulate reusable experience for fast adaptation to novel problems.

Effectively leveraging prior experience requires answering two questions: how to extract useful experience from interactions and how to apply it to new problems.
A common approach is to maintain an explicit experience library and consult it at inference time via retrieval-augmented generation (RAG) \citep{chhikara2025Mem0,fang2025memp,zhou2025memento}.
However, a key limitation is that retrieval is typically optimized for surface-level similarity rather than practical utility to the executor, and similarity does not imply effectiveness. Even when semantically close, retrieved entries may fail to yield meaningful improvements. They can miss decisive constraints or critical checks, or add noise that distracts reasoning and destabilizes solving. In addition, external libraries add noticeable inference overhead by requiring retrieval-time computation and often extra LLM calls for summarization or refinement, resulting in higher latency.

In this paper, we take a different approach: instead of maintaining a structured experience library as an external store, we encode experience in the parameters of a lightweight model and use it to guide a frozen downstream LLM executor. Accordingly, we introduce \textbf{SEAM} (\textbf{S}tructured \textbf{E}xperience \textbf{A}dapter \textbf{M}odule), a compact, trainable, and executor-specific plug-in. We view each executor as a distinct problem solver with its own strengths, inductive biases, and failure modes; SEAM therefore stores executor-tailored experience in its parameters and, at inference time, generates a structured, instance-tailored experience entry in a single forward pass, steering the frozen executor without modifying its parameters. This parameterized design eliminates the need for complex external curation and organization, since these choices are learned implicitly during training. More importantly, unlike prior RAG-style experience retrieval that relies on similarity-based selection, SEAM is optimized for utility—it is trained to generate experience entries that actually improve the frozen executor’s task performance.

We train SEAM with a forward learning procedure consisting of three training steps: (i) forward exploration, where SEAM samples multiple schema-constrained experience entries per instance; (ii) rollout-based evaluation, where a frozen executor solves each instance conditioned on each candidate and we score candidates by task success; and (iii) parametric library evolution, where we compute group-relative advantages from candidates’ returns and update SEAM under the GRPO objective while keeping the executor frozen. At test time, SEAM synthesizes a task-relevant experience entry in a single forward pass to steer the executor. Optionally, SEAM can be further improved after deployment by logging successful entries and periodically applying SFT on accumulated successes.

We conduct experiments on diverse mathematical reasoning benchmarks to assess SEAM.
Across multiple benchmarks and executors, SEAM improves accuracy with low additional overhead.
We further conduct a series of analyses to investigate SEAM’s internal mechanisms and dynamics.

\section{Related Work}
\label{sec2}
\paragraph{Learning from Experience.} Learning from experience enables LLM-based agents to reuse prior knowledge for new tasks, improving autonomy, efficiency, and performance in dynamic settings \citep{tan2025meta,zheng2023synapse,wang2024agent,liu2025contextual}. Most prior work instills this capability by directly training the agent (e.g., reinforcement learning, imitation learning, and multi-agent learning) \citep{dong2025tool,yang2024embodied,fang2025memp}, but often suffers from limited use of negative examples, weak cross-task generalization, and catastrophic forgetting. As an alternative, RAG-based frameworks extract structured experience from past rollouts, store it in a memory bank, and retrieve relevant snippets to augment prompts for new tasks \citep{suzgun2025dynamic,mishra2025can,chhikara2025Mem0}. Despite their flexibility, these pipelines rely on surface-level similarity matching that can yield low-utility prompts \citep{du2025rethinking,salama2025meminsight}, introduce nontrivial latency and compute overhead \citep{wang2024corag,quinn2025accelerating}, and depend on hand-crafted storage templates that require continual maintenance as tasks evolve \citep{xu2025mem}.

\paragraph{LLM and Reinforcement Learning.}
Reinforcement learning (RL) is increasingly used to move beyond supervised fine-tuning and enable LLMs to learn from interactive feedback. RLHF aligns outputs with human preferences \citep{ouyang2022training}, and recent work adopts GRPO to improve long-horizon reasoning and tool use under environment interaction signals. For example, Search-R1 trains models to issue web queries to maximize answer correctness \citep{jin2025search}, Tool-N1 learns when to invoke external tools \citep{zhang2025nemotron}, and RAGEN extends GRPO to multi-turn self-evolution \citep{wang2025ragen}. In learning-from-experience settings, RL is typically applied either to directly optimize the executor \citep{lu2025arpo} or to train a memory agent within RAG to extract and inject memories \citep{yan2025memory}. In contrast, SEAM applies GRPO to train a lightweight, executor-specific guidance policy that internalizes experience in parameters and generates problem-conditioned structured experience guidance to steer a frozen executor.

\begin{figure*}[!t]
    \centering
    \includegraphics[width=1.0\textwidth]{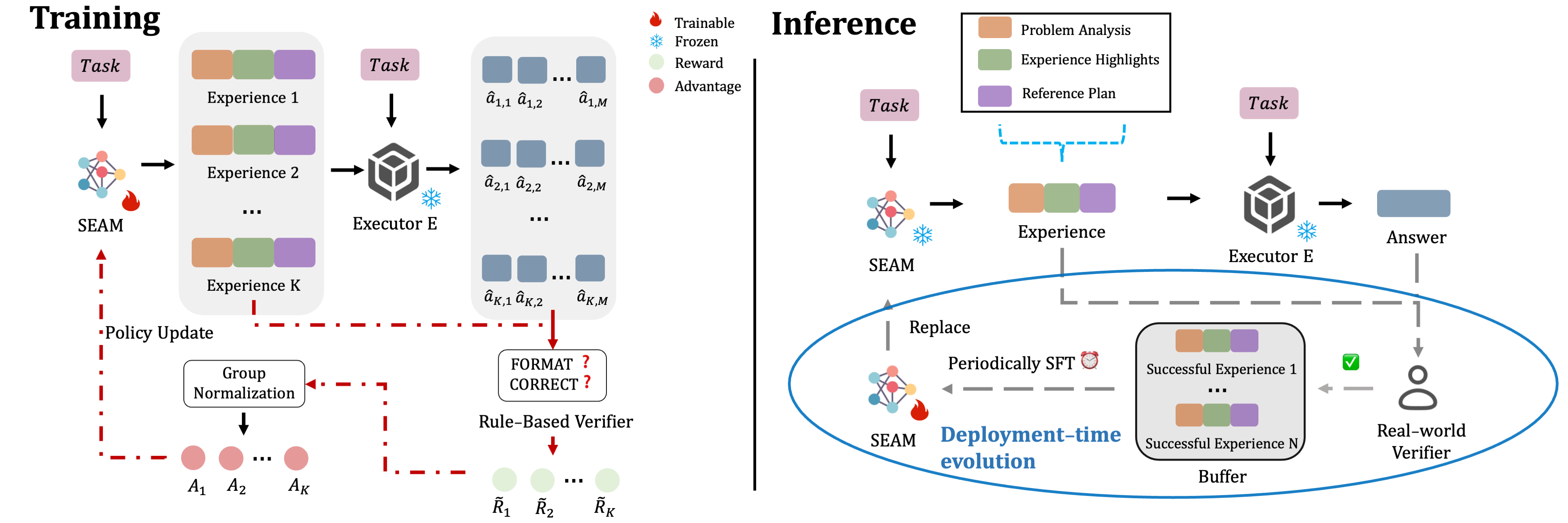} 
   \caption{Illustration of SEAM training and inference pipelines.}
    \label{fig:main}
    \vspace{-1em}
\end{figure*}
\section{Method}

\subsection{Problem Formulation}
Once deployed, LLMs operate in a fundamentally stateless manner: let $\mathcal{D}$ denote a distribution over task instances $s$, where each instance may be associated with a verifiable target $a^\star(s)$.
Given a new instance $s$ sampled from $\mathcal{D}$, a frozen executor LLM $\mathcal{E}_{\phi}$ typically generates its output $\hat{a}$ without conditioning on accumulated experience:
\begin{equation}
\hat{a} \;\sim\; \pi_{\phi}(\hat{a} \mid s).
\end{equation}
This limitation arises because LLMs are bounded by a finite context window, which prevents them from retaining
and reusing information across long interactions or evolving task sequences.
Consequently, each new instance must be reasoned through from scratch, leading to redundant exploration of
solution paths already discovered and frequent repetition of past errors.

This motivates us to equip LLMs with an experience module $\mathcal{A}_{\theta}$ that maintains experience and
provides guidance to the frozen executor. Formally, for an instance $s$ sampled from $\mathcal{D}$ and a specific executor
$\mathcal{E}_{\phi}$, the experience module produces an auxiliary structured experience prompt $z$ of limited length:
\begin{equation}
z \sim \pi_{\theta}(z \mid s, \mathcal{E}_{\phi}),
\end{equation}
The executor $\mathcal{E}_{\phi}$ then solves the task conditioned on this experience:
\begin{equation}
\hat{a} \sim \pi_{\phi}(\hat{a} \mid s, z).
\end{equation}
The design of the experience module $\mathcal{A}_{\theta}$ can thus be formulated as the following constrained
optimization problem:
\begin{equation}
\begin{array}{ll}
\displaystyle \max_{\pi_{\theta}} & \displaystyle 
\mathbb{E}_{\,s \sim \mathcal{D}, \; z \sim \pi_{\theta}(\cdot \mid s,\mathcal{E}_{\phi}), \; \hat{a} \sim \pi_{\phi}(\cdot \mid s,z)} 
\Big[ R\big(\hat{a}, a^\star(s)\big) \Big] \\
\text{s.t.} & \lVert z \rVert \leq L .
\end{array}
\end{equation}
where $R(\cdot)$ denotes a reward measuring task correctness, $a^\star(s)$ is the ground-truth (or verifiable) target for instance $s$, and $L$ is a predefined maximum length of the generated experience. In our main experiments, we instantiate $R$ using correctness on verifiable reasoning tasks; the concrete reward definition is given in Section~\ref{sec:3_3}.

Most existing work instantiates~$\mathcal{A}_{\theta}$ with the RAG pipeline: a LLM is first prompted to retrospectively
summarize its trajectories into explicit experience entries, which are stored in an external bank.
At inference time, the experience policy performs a search step that retrieves $k$ candidates from this store,
followed by a generation step that rewrites the selected items into the prompt $z$.
However, two key limitations remain: (i) \textbf{Similarity $\neq$ utility}---retrieval is optimized for surface relevance, so retrieved items can be noisy or miss decisive constraints/checks, yielding little improvement or even destabilizing the solver; (ii) \textbf{Inference overhead}---external libraries add noticeable latency due to retrieval-time computation and often extra LLM calls for refinement.

\subsection{Architecture}
To address these limitations, we adopt a fundamentally different paradigm: instead of maintaining an external experience store and retrieving entries via a RAG pipeline, we \emph{parameterize} the structured experience library by encoding it in a lightweight generator, which can be accessed with a single forward pass. We introduce \textbf{SEAM} (\textbf{S}tructured \textbf{E}xperience \textbf{A}dapter \textbf{M}odule), a compact and trainable plug-in that implicitly maintains a structured experience library in its parameters. Given a new problem instance $s$ for a downstream LLM executor $\mathcal{E}_{\phi}$, SEAM synthesizes a schema-constrained experience entry $z$ in one forward pass, and $\mathcal{E}_{\phi}$ solves the task conditioned on $z$.

This design shifts experience management away from explicit retrieval systems into a lightweight generative module and has several key characteristics.
\textbf{(1) Efficient parametric experience access.}
SEAM internalizes a structured experience library in its parameters and produces guidance via a single forward pass.
This avoids the high cost of RAG-style pipelines, notably external storage, online retrieval, and retrospective summarization by large models, yielding low overhead and low latency at inference time.
Moreover, this parameterized design eliminates the need for complex external curation and organization (e.g., indexing, clustering, deduplication, and retrieval policies), as these choices are absorbed into SEAM and learned implicitly during training.
\textbf{(2) Utility-oriented guidance instead of surface-level retrieval.}
At test time, SEAM does not retrieve entries based on embedding similarity.
Instead, it \emph{synthesizes} an instance-tailored experience entry under strict schema constraints and is trained using executor-conditioned utility signals, instantiated in our current setting by task success.
As a result, SEAM learns to generate structured experience with actionable utility for solving the current instance, rather than returning superficially relevant but low-utility snippets that may dilute or destabilize the solver.
\textbf{(3) Executor-specific experience libraries.}
We adopt an \emph{executor-specific} perspective: different downstream agents behave like different problem solvers, with distinct strengths, inductive biases, and failure modes.
Accordingly, SEAM maintains a dedicated parametric experience library for a particular frozen executor and generates \emph{dedicated}, customized, and structured guidance for that executor, without modifying the executor’s parameters.
\textbf{(4) Lightweight, pluggable modular adaptation.} We train SEAM while keeping the downstream executor $\mathcal{E}_{\phi}$ frozen, which reduces training and deployment costs, enables hot-swapping across domains, and preserves $\mathcal{E}_{\phi}$’s general competence and stability. Moreover, operational rollouts can be readily logged and recycled as feedback, enabling continual training and steady evolution of SEAM without modifying $\mathcal{E}_{\phi}$.

As illustrated in Figure~\ref{fig:main}, SEAM provides the executor with a \emph{structured experience entry} $z$ as a textual prompt. Each entry follows a fixed schema with three complementary components:
\begin{itemize}
\item \emph{Problem analysis} --- a concise assessment of the instance $s$, highlighting intrinsic difficulty, salient substructures, and likely failure modes;
\item \emph{Experience highlights} --- executor-aligned insights distilled from prior rollouts, specifying reusable strategies, heuristics, or cautionary checks that help $\mathcal{E}_{\phi}$ solve $s$;
\item \emph{Reference plan} --- an \emph{illustrative} step-by-step decomposition procedure that demonstrates a reliable workflow, rather than revealing instance-specific shortcuts or the final answer.
\end{itemize}

This schema-bound design encourages SEAM to produce guidance that is simultaneously diagnostic (problem analysis), prescriptive (experience highlights), and procedural (reference plan), providing a stable and controllable interface for steering the frozen executor. We emphasize that the schema primarily serves as a controllable interface between SEAM and the frozen executor. Its role is to constrain generated guidance into a form that can be stably and effectively exploited by a fixed solver, rather than to provide a task-specific template that is inherently well matched to a particular benchmark. In this sense, the current schema is especially natural for procedural domains, while broader domains may benefit from alternative schema designs under the same training principle.

\subsection{Training of SEAM}
\label{sec:3_3}
We train SEAM with a forward learning procedure that treats SEAM as an executor-conditioned guidance policy and improves it using observable outcomes from running a frozen executor.
SEAM is not a surrogate reasoner; it provides procedure-level, schema-constrained guidance that biases the executor’s exploration toward more reliable solution trajectories. Concretely, training iterates over three steps: (i) \textbf{forward exploration}, where SEAM samples a group of schema-constrained candidate experience entries for each instance; (ii) \textbf{rollout-based evaluation}, where we condition the frozen executor on each candidate and score its utility by downstream task success; and (iii) \textbf{policy evolution}, where we update SEAM (e.g., via GRPO) to increase the likelihood of higher-utility entries.
Throughout training, the executor remains frozen and only SEAM is updated.

\paragraph{Step 1: Forward exploration (candidate generation).}
For each training instance $s \sim \mathcal{D}$, SEAM samples a group of $K$ candidate experience entries to guide the specific executor $\mathcal{E}_{\phi}$:
\begin{equation}
\{z_j\}_{j=1}^{K} \sim \pi_{\theta}(\cdot \mid s,\mathcal{E}_{\phi}).
\label{eq:explore}
\end{equation}

\paragraph{Step 2: Rollout-based evaluation (guidance utility).}
To evaluate the guidance utility of each candidate for the target executor $\mathcal{E}_{\phi}$, we condition the frozen executor on each $z_j$ and perform $M$ stochastic rollouts.
\begin{equation}
\hat{a}_{j,m} \sim \pi_{\phi}(\cdot \mid s, z_j),
\qquad m=1,\dots,M.
\label{eq:rollout}
\end{equation}
We score each rollout using a binary reward, which jointly reflects task success and structural completeness:
\begin{equation}
\mathcal{R}(\hat{a}, z; s)=
\begin{cases}
1, & \hat{a}\text{ correct},\ z\text{ complete},\\
0, & \text{otherwise}.
\end{cases}
\label{eq:reward}
\end{equation}
Equation~\eqref{eq:reward} defines the concrete reward used in the present instantiation of SEAM. More generally, SEAM requires a utility signal for comparing multiple schema-constrained experience candidates for the same instance under a fixed executor, thereby identifying which forms of guidance are more effectively exploitable by that executor. Such a signal need not be a hard rule-based reward; depending on the setting, it may instead be instantiated using preferences, judge-based assessments, or other task-level evaluations.
We aggregate rollout outcomes by averaging rewards across rollouts for the same candidate:
\begin{equation}
\widetilde{R}_j \;=\;
\frac{1}{M}\sum_{m=1}^{M} \mathcal{R}(\hat{a}_{j,m}, z_j; s).
\label{eq:avg_reward}
\end{equation}

\paragraph{Step 3: Parametric library evolution (GRPO update).}
We update SEAM using GRPO with the group rewards $\{\widetilde{R}_j\}_{j=1}^{K}$.
Let $\pi_{\text{old}}$ denote the behavior policy used to sample $\{z_j\}$ (a snapshot of SEAM before the update),
and $\pi_{\mathrm{ref}}$ be a fixed reference policy (a copy of the initial SEAM).
We compute group-relative advantages by normalizing rewards within the group:
\begin{equation}
A_j
=\frac{\widetilde{R}_j-\bar{R}}
{\sqrt{\frac{1}{K}\sum_{i=1}^{K}\big(\widetilde{R}_i-\bar{R}\big)^2}+\delta},
\qquad
\bar{R}=\frac{1}{K}\sum_{i=1}^{K}\widetilde{R}_i,
\label{eq:adv}
\end{equation}
and perform a PPO-style clipped update over experience candidates:
\begin{align}
\mathcal{L}_{\mathrm{GRPO}}(\theta)
&= -\frac{1}{K}\sum_{j=1}^{K}
\min\!\Big(\rho_j A_j,\,
\operatorname{clip}(\rho_j,1-\epsilon,1+\epsilon)\,A_j\Big) \nonumber\\
&\quad + \beta\,
\mathrm{KL}\!\Big(
\pi_{\theta}(\cdot\mid s,\mathcal{E}_{\phi})
\,\big\|\,
\pi_{\mathrm{ref}}(\cdot\mid s,\mathcal{E}_{\phi})
\Big), \nonumber\\
\rho_j
&=\frac{\pi_{\theta}(z_j\mid s,\mathcal{E}_{\phi})}
{\pi_{\text{old}}(z_j\mid s,\mathcal{E}_{\phi})}.
\label{eq:grpo}
\end{align}
Crucially, gradients are backpropagated only through SEAM; the executor $\mathcal{E}_{\phi}$ remains frozen. From a high-level perspective, we optimize one model using the observable execution behavior of another: we treat $\mathcal{E}_{\phi}$ as a fixed solver that provides behavioral signals, and train SEAM to generate the structured guidance that $\mathcal{E}_{\phi}$ can best exploit. In this way, SEAM is not a surrogate reasoner; instead, it supplies procedure-level, schema-constrained cues that bias the executor’s exploration toward more reliable solution trajectories. This improves guidance and end-to-end effectiveness without backpropagating through $\mathcal{E}_{\phi}$. Such a decoupled paradigm enables SEAM to continually improve from execution feedback without modifying $\mathcal{E}_{\phi}$’s parameters, thereby preserving the executor’s general competence while enhancing robustness and overall performance.

\paragraph{Optional: Deployment-time evolution via logged SFT.}
As an optional extension, SEAM can further improve during deployment by logging successful experience entries that actually guide the frozen executor to correct solutions. When enabled, we store these pairs $(s, z^{\star})$ in a buffer $\mathcal{B}$ and periodically fine-tune SEAM with teacher forcing:
\begin{align}
\mathcal{L}_{\mathrm{SFT}}(\theta)
&= - \mathbb{E}_{(s, z^{\star}) \sim \mathcal{B}}
\Big[\log \pi_{\theta}(z^{\star} \mid s, \mathcal{E}_{\phi})\Big].
\label{eq:sft}
\end{align}
This update again modifies only SEAM while keeping the executor frozen, progressively internalizing effective experience from concrete past episodes into the parametric library.

\section{Experiment}
\label{sec:experiments}
\subsection{Experiment Settings}

\paragraph{Datasets.} 
For training, we randomly sample 5k instances from the \textsc{DAPO} dataset, which provides diverse problem statements with annotated solutions.
Evaluation is conducted on four widely used mathematical reasoning benchmarks, including GSM8K \citep{cobbe2021gsm8k}, MATH \citep{hendrycks2021measuring}, AIME24 and AIME25.
These benchmarks cover both routine arithmetic and competition-level mathematics, enabling us to assess generalization across varying difficulty.
We use pass@1 as the primary metric and report the mean over 10 independent runs per dataset.
In addition, for our cross-domain analysis on code and QA benchmarks including CodeContests \citep{li2022alphacode}, MBPP \citep{austin2021program}, HotpotQA \citep{yang-etal-2018-hotpotqa}, and Natural Questions (NQ) \citep{kwiatkowski-etal-2019-natural}, we evaluate on a random subset of 300 instances from each benchmark.\footnote{Dataset details and access links are provided in \S~\ref{app:datasets}.}

\paragraph{Baselines.}
In our main experiments, we compare SEAM with several representative baselines.
Together, these baselines span three families: (i) direct use of the original model without experience memory;
(ii) directly training the executor, which can be viewed as enabling the executor to internalize and apply experience;
and (iii) RAG-based approaches for learning from experience.

\begin{itemize}[leftmargin=*, itemsep=0.25em, topsep=0.25em]
    \item \textbf{Original Model.}
    The frozen executor $\mathcal{E}_{\phi}$ is directly applied to test problems without any memory augmentation
    or additional training. This serves as the minimal baseline.

    \item \textbf{Training-based method.} Directly train the executor $\mathcal{E}_{\phi}$ using GRPO, baking prior experience
    into $\mathcal{E}_{\phi}$'s parameters.

    \item \textbf{RAG-based methods.}
    \begin{itemize}[leftmargin=1.2em, itemsep=0.2em, topsep=0.2em]
        \item \textbf{MEM-0} \citep{chhikara2025Mem0}:
        A scalable long-term memory framework that dynamically extracts--integrates--retrieves salient information to form a persistent external memory for relational reasoning.

        \item \textbf{Dynamic-Cheatsheet (DC)} \citep{suzgun2025dynamic}:
        Augments black-box LMs with a persistent, self-curated test-time memory of concise, transferable strategies.
        We adopt the DC-RS (retrieve-and-synthesize) variant.

        \item \textbf{Memento} \citep{zhou2025memento}:
        Equips black-box LLMs with an external episodic memory, retrieving and adapting prior trajectories for case-based planning without fine-tuning; supports learned Q-function retrieval with online write/read.
    \end{itemize}
\end{itemize}

\paragraph{Implementation Details.}
We adopt \textsc{Qwen3-0.6B}~\citep{yang2025qwen3} as the base model for SEAM and, unless otherwise specified,
report results with the GRPO-trained SEAM (without deployment-time SFT).
We use \textsc{Qwen3-4B}~\citep{yang2025qwen3} and \textsc{Deepseek-R1-distilled-LLaMA-3.1-8B}~\citep{guo2025deepseek} as executors in the main experiments,
and all baselines share the same frozen executor (i.e., their \emph{executor} is identical to ours).
We enable \textsc{Qwen3} thinking mode throughout.
Unless otherwise specified, we cap the maximum length of each generated experience entry to 4k tokens and the executor output to 8k tokens.
For RAG-based baselines, we run them on top of the same frozen executor; for methods with an external experience bank, we pre-fill the bank using experiences constructed from the same training set, and then update it sequentially after each evaluation instance.
All methods use greedy decoding for evaluation, and all experiments are conducted on 4 NVIDIA A100 (80GB) GPUs.\footnote{More details are provided in \S~\ref{app:detail}.}

\subsection{Main Results}
\label{exp:main}
Table~\ref{tab:main} shows across four math-reasoning benchmarks and two executors, SEAM achieves the best performance on all datasets, demonstrating the effectiveness and generality across different executors.
Relative to directly training the executor, SEAM attains superior overall performance, indicating that internalizing the experience in a small parameterized generator that provides task-conditioned test-time guidance is effective and can even surpass directly optimizing the executor itself. This likely reflects an inductive-bias difference: directly training the executor updates a global behavior policy and aggregates credit across heterogeneous instances, whereas SEAM learns instance-conditioned guidance that injects procedural priors and reshapes the executor’s search space on a per-problem basis without perturbing the executor’s general reasoning capabilities.

Compared with RAG-based approaches, SEAM delivers substantially stronger performance, indicating that its mechanism of implicitly parameterizing experience and directly generating task-specific guidance addresses key limitations of RAG such as low-quality experience summarization and weak downstream guidance. 
Regarding task difficulty, the largest performance gains of SEAM emerge on the more challenging AIME problems, whereas on the comparatively saturated MATH500 and GSM8K benchmarks, SEAM still provides modest improvements. This pattern suggests that experience-based guidance becomes increasingly critical and effective for the executor when tackling harder tasks. Collectively, these findings affirm the robustness and effectiveness of our design across diverse models and datasets.

\begin{table*}[ht]
\centering
\small
\caption{Main results across four math-reasoning benchmarks, evaluated by pass@1.}
\begin{tabular}{@{}lcccc@{}}
\toprule
\textbf{Methods} & \multicolumn{1}{c}{\textbf{AIME24}} & \multicolumn{1}{c}{\textbf{AIME25}} & \multicolumn{1}{c}{\textbf{MATH500}} & \multicolumn{1}{c}{\textbf{GSM8K}}\\
\midrule
\midrule
{\textbf{\textsc{Qwen3-4B}}} \\
\midrule
Original Model & 0.407 & 0.380 & 0.784 & 0.932\\
\midrule
GRPO  & 0.473 & 0.440 &  0.756 & 0.923 \\
\midrule
MEM-0 & 0.413 & 0.403 & 0.792 & 0.938 \\
Dynamic-Cheatsheet & 0.447 & 0.417 & 0.803 & 0.943 \\
Memento & 0.463 & 0.433 & 0.811 & 0.947 \\
\midrule
SEAM & \textbf{0.504} & \textbf{0.460}  & \textbf{0.818} & \textbf{0.958} \\
\midrule
\midrule

{\textbf{\textsc{Deepseek-R1-distilled-LLaMA-3.1-8B}}} \\
\midrule
Original Model & 0.357 & 0.243 & 0.734 & 0.773 \\
\midrule
GRPO  & 0.437 & 0.323 & 0.795 & 0.852 \\
\midrule
MEM-0 & 0.397 & 0.297 & 0.768 & 0.812 \\
Dynamic-Cheatsheet & 0.383 & 0.283 & 0.774 & 0.817 \\
Memento & 0.413 & 0.320 & 0.787 & 0.829 \\
\midrule
SEAM  & \textbf{0.442} & \textbf{0.352}  & \textbf{0.794} & \textbf{0.856} \\
\bottomrule
\end{tabular}
\label{tab:main}
\end{table*}

\subsection{Ablation Study}
\begin{table}[t]
\centering
\small
\caption{Ablation on training regimes with the \textsc{Qwen3-4B} executor (pass@1; mean over 10 random $\mathcal{D}_{\mathrm{eval}}$ splits).}
\resizebox{\columnwidth}{!}{\begin{tabular}{lcc}
\midrule
\textbf{Methods} & \textbf{AIME25} & \textbf{MATH500} \\
\midrule
\textit{SEAM} & 0.460 & 0.818 \\
\midrule
\textit{SEAM without training} & 0.374 & 0.790 \\
\textit{SEAM with SFT only} & 0.425 & 0.809 \\
\textit{SEAM with deployment-time evolution} & \textbf{0.482} & \textbf{0.836} \\
\midrule
\end{tabular}}
\label{tab:ablation_regimes}
\end{table}

\paragraph{Training regimes.}
We compare four training regimes:
(i) \textit{SEAM without training}, which uses the same prompting strategy and experience format but applies no learning;
(ii) \textit{SEAM}, trained on the sampled DAPO dataset using GRPO;
(iii) \textit{SEAM with SFT-only}, trained only with online teacher forcing on logged successful data; and
(iv) \textit{SEAM with deployment-time evolution}, which after GRPO training applies the same logged successful data.

For each benchmark, we randomly split the test set into a 30\% stream $\mathcal{D}_{\mathrm{stream}}$ and a disjoint 70\% evaluation set $\mathcal{D}_{\mathrm{eval}}$.
To collect data for online SFT, we keep the executor frozen and process instances from $\mathcal{D}_{\mathrm{stream}}$.
Whenever SEAM’s experience leads the executor to a correct solution, we log the successful pair $(s, z^{\star})$ into a buffer.
We iterate over $\mathcal{D}_{\mathrm{stream}}$ for up to 10 rounds, stopping early once the buffer reaches 1k successful samples or when 10 rounds are completed.


\paragraph{Results and discussion.}
Table~\ref{tab:ablation_regimes} shows that training is essential beyond prompting and experience formatting.
SFT-only provides a modest improvement over no training, but remains less effective than GRPO, suggesting that directly maximizing the likelihood of logged successful experiences is insufficient to reliably optimize guidance utility under a frozen executor.
In contrast, GRPO-trained SEAM learns to generate structured experiences that are better aligned with both the task and the frozen executor’s execution behavior, yielding more informative problem analyses and actionable step-wise strategies, and thereby providing consistent improvements.
Finally, deployment-time evolution offers an additional refinement by internalizing experiences that demonstrably succeed in situ, helping SEAM adapt to deployment-specific patterns and distributional shifts.
More principled deployment-time update schemes remain an interesting direction for future work.

\section{Analysis}
To better understand SEAM, we analyze its efficiency, robustness to experience size, and generalization across model pairings and domains.

\subsection{Efficiency}

\begin{table}[t]
\centering
\small
\caption{Efficiency comparison on \textsc{DeepSeek-R1-Distilled-Llama-3.1-8B} using \(4\times\)A100\,80GB, evaluated on \textsc{AIME25}. GRPO (Full) denotes full-parameter GRPO training, while GRPO (LoRA) denotes GRPO training with LoRA adapters. 
Training efficiency is reported as GPU-hours to convergence. 
Inference efficiency is measured by the average end-to-end wall-clock latency per problem (\textit{Per-inst.}) and by \textit{Time-to-correct (TTC)}, defined as the total inference time over the evaluation set divided by the number of correctly solved problems.}
\setlength{\tabcolsep}{3pt}
\begin{tabularx}{\columnwidth}{@{}lccc@{}}
\toprule
\textbf{Method} &
\shortstack{\textbf{Training}\\\textbf{Efficiency}} &
\multicolumn{2}{c}{\textbf{Inference Efficiency}} \\
\cmidrule(lr){3-4}
& \shortstack{\textbf{GPU-hours}\\\textbf{to conv.}} &
\shortstack{\textbf{Per-inst.}\\\textbf{(s)}} &
\shortstack{\textbf{TTC}\\\textbf{(s)}} \\
\midrule
Original Model      & --  & 0.85 & 3.50 \\
GRPO (Full)                & 104 & \textbf{0.81} & \textbf{2.51} \\
GRPO (LoRA)                & 104 & \textbf{0.81} & \textbf{2.51} \\
MEM-0               & --  & 3.86 & 13.00 \\
Dynamic-Cheatsheet  & --  & 4.05 & 14.31 \\
Memento             & --  & 3.93 & 12.28 \\
\textbf{SEAM}      & \textbf{53}  &0.95 & 2.70 \\
\bottomrule
\end{tabularx}
\label{tab:eff}
\end{table}

Table~\ref{tab:eff} summarizes efficiency along three axes: training efficiency, per-instance inference latency, and time-to-correct\footnote{We run the system once per instance. Let $N$ be the number of evaluation instances, $t_i$ the end-to-end wall-clock latency for instance $i$, and $c_i=\mathbb{I}[\hat{a}_i \text{ is correct}]$ the correctness indicator. Then
$\mathrm{TTC}=\frac{\sum_{i=1}^{N} t_i}{\sum_{i=1}^{N} c_i}
=\frac{\bar{t}}{\mathrm{Acc}}$, where $\bar{t}=\frac{1}{N}\sum_{i=1}^{N} t_i$ and $\mathrm{Acc}=\frac{1}{N}\sum_{i=1}^{N} c_i$.}. During training, our method is substantially more efficient than directly fine-tuning a large model, because only the lightweight module is optimized while the executor remains frozen, which reduces compute requirements. At inference time, SEAM introduces a small, fixed, per-instance latency due to generating a structured experience that augments the executor. When measured by time-to-correct, SEAM exhibits clear advantages: the auxiliary experience it provides raises task success rates and shortens average problem-solving time compared with RAG-based baselines. Overall, the design delivers notable gains in both training efficiency and solution speed while maintaining competitive inference latency.

\begin{table}[t]
\centering
\small
\caption{AIME24 pass@1 for pairing SEAM scales with executors. All SEAM models are trained independently with rollouts from their respective paired executors; \textit{w/o SEAM} reports the executor-only baseline without any extra training.}
\begin{tabular}{@{}lccc@{}}
\toprule
& \multicolumn{3}{c}{\textbf{Executor}}\\
\cmidrule(lr){2-4}
\textbf{SEAM} & \textbf{Qwen3-4B} & \textbf{Qwen3-8B} & \textbf{Avg.} \\
\midrule
\textit{w/o SEAM} & 0.407 & 0.635 & 0.521 \\
\midrule
SEAM-Qwen3-0.6B & 0.504 & 0.643 & 0.574 \\
SEAM-Qwen3-1.7B & \textbf{0.517} & \textbf{0.697} & \textbf{0.607} \\
\bottomrule
\end{tabular}
\label{tab:pairing}
\end{table}

\subsection{Model Pairing}
We systematically examine the pairings of SEAM’s different lightweight models with various executors in Table~\ref{tab:pairing}.
On the side of smaller lightweight models, Qwen3-0.6B yields some improvements, but its limited generalization and generation quality lead to experience entries that lack stability and coverage in complex tasks, resulting in only modest gains. In contrast, Qwen3-1.7B exhibits stronger generalization and more structured generation, thereby providing more stable and significant improvements for the executor on harder tasks. On the side of executors, stronger LLMs are already capable of solving a substantial portion of problems correctly on their own, which reduces the marginal benefit brought by external experience, making the improvements from SEAM less pronounced. Overall, there exists a complementary relationship between the scale of SEAM and the capability of the executor: given a fixed executor, enhancing the generative capacity of SEAM effectively amplifies the value of external experience; whereas when the executor itself is already sufficiently strong, the gains from external experience saturate, in which case further strengthening SEAM (e.g., with a more powerful base model or improved training) may be necessary to achieve additional benefits.

\begin{table}[t]
\centering
\small
\caption{Cross-executor transfer on \textsc{AIME25}: relative improvement (\%) when applying a SEAM trained with \textsc{Qwen3-4B} as the executor to other executors, without any additional training. Relative improvement is computed as \(\big[(\text{with SEAM}-\text{original})/\text{original}\big]\times 100\%\).}
\setlength{\tabcolsep}{6pt}
\begin{tabularx}{\columnwidth}{@{}l c@{}}
\toprule
\textbf{Target Executor} & \textbf{Rel. imp.} \\
\midrule
\textsc{DeepSeek-R1-Distilled-Llama-3.1-8B} & +27.9\% \\
\textsc{GPT-4o} & +22.7\% \\
\textsc{Claude-3.5-Sonnet} & +25.3\% \\
\bottomrule
\end{tabularx}
\label{tab:cross_model}
\end{table}

\subsubsection{Cross-model Generalization}
\label{sec:5_3_1}
We further conducted a cross-executor transfer experiment to examine the portability of SEAM across different executors.
Specifically, we directly migrated the SEAM trained in \S\ref{exp:main} with \textsc{Qwen3-4B} as the executor to other executors of different architecture or scale, without any additional training or adaptation. The results in Table~\ref{tab:cross_model} show that the transferred SEAM can still provide a certain degree of performance improvement on the new executor, indicating that the generated experience exhibits some level of generality. However, for \textsc{Deepseek-R1-distilled-LLaMA-3.1-8B} executor, the improvement is relatively limited and remains below the performance of the original SEAM–executor pairing (44.8\%\footnote{The result is computed from the data reported in Table \ref{tab:main}.} vs. 27.9\%). This limitation arises from SEAM being optimized for a specific executor; variations in executors’ capabilities and characteristics hinder portability and can degrade performance under direct transfer. To address this, future work could explore joint training across multiple executors.

\subsubsection{Cross-domain Generalization}
\begin{figure}[!t]
    \centering
    \includegraphics[width=0.49\textwidth]{graph/seam_barplot_aclstyle_warm.png} 
   \caption{The cross-domain results for SEAM with Qwen3-4B as executor.}
    \label{fig:cross}
\end{figure}
We also investigated the cross-domain capability of SEAM. We directly use the SEAM trained in \S\ref{exp:main} with \textsc{Qwen3-4B} as the executor as well. As illustrated in Figure~\ref{fig:cross}, although SEAM is trained solely on mathematical data, it still enhances performance in two markedly different domains. This suggests that SEAM acquires not just math-specific skills but a more general reasoning framework, including problem decomposition, explicit step planning, and the avoidance of common pitfalls. When the executor already possesses a baseline level of competence, these structured cues further improve its effectiveness. However, the cross-domain gains are relatively limited, because SEAM lacks experience for these unfamiliar domains. Enabling a single SEAM to efficiently switch across and adapt to different domains is a key direction for future.

\subsection{Experience Scaling}
\begin{figure}[t]
  \centering
  \begin{subfigure}[t]{0.49\columnwidth}
    \centering
    \includegraphics[width=\linewidth]{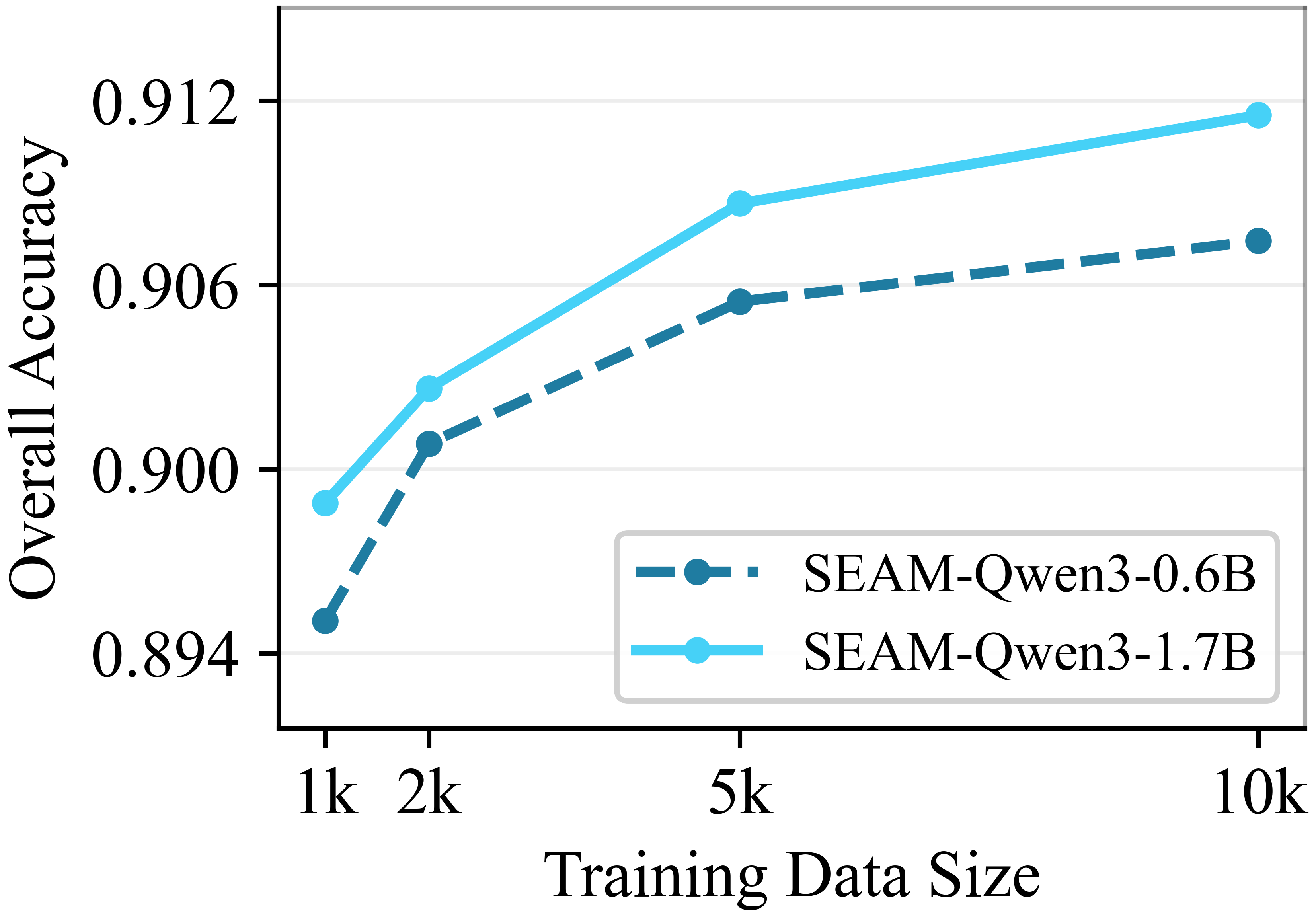}
    \caption{\textsc{Qwen3-4B} executor.}
    \label{fig:exp-qwen}
  \end{subfigure}\hfill
  \begin{subfigure}[t]{0.49\columnwidth}
    \centering
    \includegraphics[width=\linewidth]{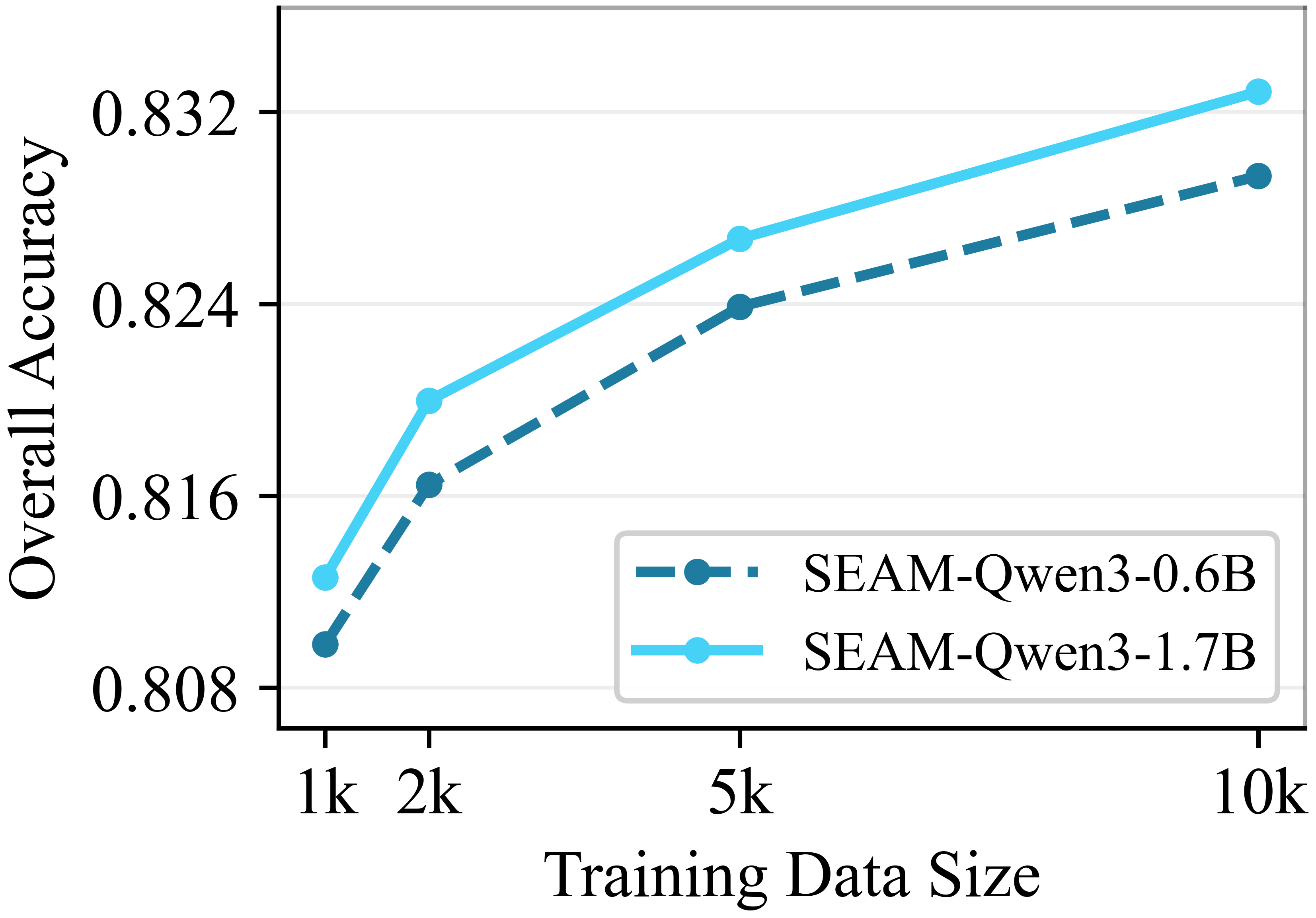}
      \caption{\textsc{Deepseek-R1} executor.}

    \label{fig:exp-deepseek}
  \end{subfigure}
  \caption{The results of scaling with experience size.}
  \label{fig:exp-scaling}
\end{figure}

We study the effect of experience size by changing the number of entries used to train SEAM while keeping the executor fixed, and evaluating overall accuracy aggregated over all instances from the four mathematical benchmarks (Figure~\ref{fig:exp-scaling}). Performance improves monotonically as the experience size grows, indicating that SEAM benefits primarily from accumulating and reusing more high-quality experiences rather than from sampling luck. The gains show diminishing returns at larger scales, with improvements gradually tapering off, suggesting that further progress likely requires richer diversity and better coverage of experience. Larger SEAM variants consistently outperform smaller ones and gain more from additional experiences, revealing a complementary relationship between model capacity and experience scale. These trends are consistent across executors, demonstrating robust and transferable scaling behavior.

\section{Conclusion}
We presented \textbf{SEAM}, a lightweight, executor-specific module that parameterizes experience and produces structured, problem-conditioned guidance in a single forward pass to steer a frozen LLM executor. SEAM is trained with a three-step pipeline: (1) generate multiple schema-constrained experience candidates; (2) evaluate each candidate via executor rollouts and task success; and (3) update SEAM with GRPO using group-relative advantages to internalize the most effective experiences.
SEAM can further evolve at deployment time by applying SFT on accumulated successes. Experiments and additional analyses demonstrate SEAM’s effectiveness and robustness across different frozen executors, and position it as a practical alternative to retrieval-based experience reuse. We hope this work motivates a broader paradigm of learning from experience beyond RAG.

\section*{Limitations}
Despite the promising results, our current evidence remains subject to several important limitations.

\paragraph{Limited scope of current evidence.}
Our empirical evaluation is primarily concentrated on mathematical reasoning benchmarks and a small set of executors. While these results support the effectiveness of SEAM in this setting, they do not yet establish equally strong conclusions for broader task families, more diverse domains, or substantially different executor classes.

\paragraph{Executor-specificity as a design trade-off.}
SEAM is deliberately trained in an executor- and domain-specific manner so that the generated guidance matches a fixed solver's strengths, reasoning preferences, and failure modes. As a result, a single SEAM instance that adaptively transfers across heterogeneous models and domains is not a demonstrated property of the current system, but rather an open problem for future work.

\paragraph{Continual evolution remains underexplored.}
Although SEAM can be further updated using logged feedback, we do not yet study its robustness under distribution shift, noisy or delayed supervision, and long-term continual adaptation. In particular, reliable experience evolution over time without degradation or catastrophic forgetting remains insufficiently characterized in the current work.
\section*{Ethical Considerations}
Our work studies how to improve a frozen large-language executor by training a lightweight guidance model (SEAM) from execution feedback. We discuss potential ethical risks and mitigations below.

\paragraph{Intended use and misuse.}
SEAM is designed to generate structured, task-conditioned guidance that improves reliability and efficiency for benign problem-solving tasks (e.g., mathematical reasoning).
Like other LLM-based systems, however, it could be repurposed to assist with harmful activities by providing step-by-step procedural guidance.
We do not propose SEAM as a safety mechanism, and our method should be deployed with standard safeguards (e.g., content filtering, policy enforcement, and monitoring) when used in real-world applications.

\paragraph{Data and privacy.}
Our experiments use public datasets and do not involve personal data collection.
In the optional deployment-time evolution setting, the system may log successful interactions for supervised fine-tuning.
If applied beyond our experimental setup, such logging could capture user-provided sensitive information.
Deployments should therefore minimize data retention, anonymize or redact sensitive fields, and obtain appropriate user consent; access controls and secure storage should be used for any retained logs.

\paragraph{Bias and fairness.}
SEAM internalizes patterns from the executor's rollouts and the training data distribution.
As a result, any biases present in the executor or datasets may be propagated or amplified in generated guidance.
While our benchmarks are primarily technical, future applications in open-domain settings should include bias assessment and mitigation, such as evaluating guidance quality across user groups and enforcing constraints on unsafe or discriminatory content.

\paragraph{Robustness and over-reliance.}
Generated guidance may be incorrect, incomplete, or misleading, which could cause downstream errors if users or systems rely on it uncritically.
Our design partially mitigates this by keeping the executor frozen and by training SEAM to optimize downstream task success, but this does not guarantee correctness on all inputs.
For higher-stakes deployments, additional measures (e.g., verification, calibration, or human oversight) are necessary.

\paragraph{Environmental impact.}
Training SEAM requires additional computation beyond using a fixed executor alone.
However, SEAM is lightweight compared to the executor and can reduce time-to-correct at inference by improving solution success rates.
We report efficiency metrics to support transparent accounting of computational costs.

\paragraph{Responsible release.}
We release our code in the supplementary material and recommend accompanying it with clear documentation on intended use, limitations, and safety considerations, as well as default configurations that disable deployment-time logging unless explicitly enabled by users.

\section*{Acknowledgments}
This work was supported by the Research Project of Quancheng Laboratory, China (Grant No. QCL20250105), and the Key R\&D Program of Shandong Province (Grant No. SYS202201).


\bibliography{main}

@article{suzgun2025dynamic,
  title={Dynamic cheatsheet: Test-time learning with adaptive memory},
  author={Suzgun, Mirac and Yuksekgonul, Mert and Bianchi, Federico and Jurafsky, Dan and Zou, James},
  journal={arXiv preprint arXiv:2504.07952},
  year={2025}
}

@article{cobbe2021gsm8k,
  title={Training Verifiers to Solve Math Word Problems},
  author={Cobbe, Karl and Kosaraju, Vineet and Bavarian, Mohammad and Chen, Mark and Jun, Heewoo and Kaiser, Lukasz and Plappert, Matthias and Tworek, Jerry and Hilton, Jacob and Nakano, Reiichiro and Hesse, Christopher and Schulman, John},
  journal={arXiv preprint arXiv:2110.14168},
  year={2021}
}

@article{hendrycks2021measuring,
  title={Measuring mathematical problem solving with the math dataset},
  author={Hendrycks, Dan and Burns, Collin and Kadavath, Saurav and Arora, Akul and Basart, Steven and Tang, Eric and Song, Dawn and Steinhardt, Jacob},
  journal={arXiv preprint arXiv:2103.03874},
  year={2021}
}

@article{mishra2025can,
  title={How Can Input Reformulation Improve Tool Usage Accuracy in a Complex Dynamic Environment? A Study on $\tau$-bench},
  author={Mishra, Venkatesh and Saeidi, Amir and Raj, Satyam and Nakamura, Mutsumi and Srinivasa, Jayanth and Liu, Gaowen and Payani, Ali and Baral, Chitta},
  journal={arXiv preprint arXiv:2508.20931},
  year={2025}
}

@article{chhikara2025mem0,
  title={Mem0: Building production-ready ai agents with scalable long-term memory},
  author={Chhikara, Prateek and Khant, Dev and Aryan, Saket and Singh, Taranjeet and Yadav, Deshraj},
  journal={arXiv preprint arXiv:2504.19413},
  year={2025}
}

@article{ahn2024large,
  title={Large language models for mathematical reasoning: Progresses and challenges},
  author={Ahn, Janice and Verma, Rishu and Lou, Renze and Liu, Di and Zhang, Rui and Yin, Wenpeng},
  journal={arXiv preprint arXiv:2402.00157},
  year={2024}
}

@article{naveed2025comprehensive,
  title={A comprehensive overview of large language models},
  author={Naveed, Humza and Khan, Asad Ullah and Qiu, Shi and Saqib, Muhammad and Anwar, Saeed and Usman, Muhammad and Akhtar, Naveed and Barnes, Nick and Mian, Ajmal},
  journal={ACM Transactions on Intelligent Systems and Technology},
  volume={16},
  number={5},
  pages={1--72},
  year={2025},
  publisher={ACM New York, NY}
}

@article{guo2025deepseek,
  title={Deepseek-r1: Incentivizing reasoning capability in llms via reinforcement learning},
  author={Guo, Daya and Yang, Dejian and Zhang, Haowei and Song, Junxiao and Zhang, Ruoyu and Xu, Runxin and Zhu, Qihao and Ma, Shirong and Wang, Peiyi and Bi, Xiao and others},
  journal={arXiv preprint arXiv:2501.12948},
  year={2025}
}

@article{fan2025if,
  title={If an LLM Were a Character, Would It Know Its Own Story? Evaluating Lifelong Learning in LLMs},
  author={Fan, Siqi and Huang, Xiusheng and Yao, Yiqun and Fang, Xuezhi and Liu, Kang and Han, Peng and Shang, Shuo and Sun, Aixin and Wang, Yequan},
  journal={arXiv preprint arXiv:2503.23514},
  year={2025}
}

@inproceedings{yu2025stateful,
  title={Stateful large language model serving with pensieve},
  author={Yu, Lingfan and Lin, Jinkun and Li, Jinyang},
  booktitle={Proceedings of the Twentieth European Conference on Computer Systems},
  pages={144--158},
  year={2025}
}

@article{feng2025get,
  title={Get Experience from Practice: LLM Agents with Record \& Replay},
  author={Feng, Erhu and Zhou, Wenbo and Liu, Zibin and Chen, Le and Dong, Yunpeng and Zhang, Cheng and Zhao, Yisheng and Du, Dong and Hua, Zhichao and Xia, Yubin and others},
  journal={arXiv preprint arXiv:2505.17716},
  year={2025}
}

@article{efklides2009role,
  title={The role of metacognitive experiences in the learning process},
  author={Efklides, Anastasia},
  journal={Psicothema},
  pages={76--82},
  year={2009}
}

@book{kolb2014experiential,
  title={Experiential learning: Experience as the source of learning and development},
  author={Kolb, David A},
  year={2014},
  publisher={FT press}
}

@article{usher1993experiential,
  title={Experiential learning or learning from experience: Does it make a difference},
  author={Usher, Robin},
  journal={Using experience for learning},
  pages={169--180},
  year={1993},
  publisher={Society for Research into Higher Education and Open University Press Bristol, UK}
}

@article{fang2025memp,
  title={Memp: Exploring agent procedural memory},
  author={Fang, Runnan and Liang, Yuan and Wang, Xiaobin and Wu, Jialong and Qiao, Shuofei and Xie, Pengjun and Huang, Fei and Chen, Huajun and Zhang, Ningyu},
  journal={arXiv preprint arXiv:2508.06433},
  year={2025}
}

@article{zhou2025memento,
  title={Memento: Fine-tuning llm agents without fine-tuning llms},
  author={Zhou, Huichi and Chen, Yihang and Guo, Siyuan and Yan, Xue and Lee, Kin Hei and Wang, Zihan and Lee, Ka Yiu and Zhang, Guchun and Shao, Kun and Yang, Linyi and others},
  journal={Preprint},
  year={2025}
}

@article{du2025rethinking,
  title={Rethinking memory in ai: Taxonomy, operations, topics, and future directions},
  author={Du, Yiming and Huang, Wenyu and Zheng, Danna and Wang, Zhaowei and Montella, Sebastien and Lapata, Mirella and Wong, Kam-Fai and Pan, Jeff Z},
  journal={arXiv preprint arXiv:2505.00675},
  year={2025}
}

@article{wang2024corag,
  title={Corag: A cost-constrained retrieval optimization system for retrieval-augmented generation},
  author={Wang, Ziting and Yuan, Haitao and Dong, Wei and Cong, Gao and Li, Feifei},
  journal={arXiv preprint arXiv:2411.00744},
  year={2024}
}

@inproceedings{quinn2025accelerating,
  title={Accelerating retrieval-augmented generation},
  author={Quinn, Derrick and Nouri, Mohammad and Patel, Neel and Salihu, John and Salemi, Alireza and Lee, Sukhan and Zamani, Hamed and Alian, Mohammad},
  booktitle={Proceedings of the 30th ACM International Conference on Architectural Support for Programming Languages and Operating Systems, Volume 1},
  pages={15--32},
  year={2025}
}

@article{salama2025meminsight,
  title={Meminsight: Autonomous memory augmentation for llm agents},
  author={Salama, Rana and Cai, Jason and Yuan, Michelle and Currey, Anna and Sunkara, Monica and Zhang, Yi and Benajiba, Yassine},
  journal={arXiv preprint arXiv:2503.21760},
  year={2025}
}

@article{xu2025mem,
  title={A-mem: Agentic memory for llm agents},
  author={Xu, Wujiang and Mei, Kai and Gao, Hang and Tan, Juntao and Liang, Zujie and Zhang, Yongfeng},
  journal={arXiv preprint arXiv:2502.12110},
  year={2025}
}

@inproceedings{tan2025meta,
  title={Meta-Agent-Workflow: Streamlining Tool Usage in LLMs through Workflow Construction, Retrieval, and Refinement},
  author={Tan, Xiaoyu and Li, Bin and Qiu, Xihe and Qu, Chao and Chu, Wei and Xu, Yinghui and Qi, Yuan},
  booktitle={Companion Proceedings of the ACM on Web Conference 2025},
  pages={458--467},
  year={2025}
}

@article{zheng2023synapse,
  title={Synapse: Trajectory-as-exemplar prompting with memory for computer control},
  author={Zheng, Longtao and Wang, Rundong and Wang, Xinrun and An, Bo},
  journal={arXiv preprint arXiv:2306.07863},
  year={2023}
}

@article{wang2024agent,
  title={Agent workflow memory},
  author={Wang, Zora Zhiruo and Mao, Jiayuan and Fried, Daniel and Neubig, Graham},
  journal={arXiv preprint arXiv:2409.07429},
  year={2024}
}

@article{liu2025contextual,
  title={Contextual Experience Replay for Self-Improvement of Language Agents},
  author={Liu, Yitao and Si, Chenglei and Narasimhan, Karthik and Yao, Shunyu},
  journal={arXiv preprint arXiv:2506.06698},
  year={2025}
}

@article{dong2025tool,
  title={Tool-Star: Empowering LLM-Brained Multi-Tool Reasoner via Reinforcement Learning},
  author={Dong, Guanting and Chen, Yifei and Li, Xiaoxi and Jin, Jiajie and Qian, Hongjin and Zhu, Yutao and Mao, Hangyu and Zhou, Guorui and Dou, Zhicheng and Wen, Ji-Rong},
  journal={arXiv preprint arXiv:2505.16410},
  year={2025}
}

@inproceedings{yang2024embodied,
  title={Embodied multi-modal agent trained by an llm from a parallel textworld},
  author={Yang, Yijun and Zhou, Tianyi and Li, Kanxue and Tao, Dapeng and Li, Lusong and Shen, Li and He, Xiaodong and Jiang, Jing and Shi, Yuhui},
  booktitle={Proceedings of the IEEE/CVF conference on computer vision and pattern recognition},
  pages={26275--26285},
  year={2024}
}

@article{ouyang2022training,
  title={Training language models to follow instructions with human feedback},
  author={Ouyang, Long and Wu, Jeffrey and Jiang, Xu and Almeida, Diogo and Wainwright, Carroll and Mishkin, Pamela and Zhang, Chong and Agarwal, Sandhini and Slama, Katarina and Ray, Alex and others},
  journal={Advances in neural information processing systems},
  volume={35},
  pages={27730--27744},
  year={2022}
}

@article{jin2025search,
  title={Search-r1: Training llms to reason and leverage search engines with reinforcement learning},
  author={Jin, Bowen and Zeng, Hansi and Yue, Zhenrui and Yoon, Jinsung and Arik, Sercan and Wang, Dong and Zamani, Hamed and Han, Jiawei},
  journal={arXiv preprint arXiv:2503.09516},
  year={2025}
}

@article{zhang2025nemotron,
  title={Nemotron-research-tool-n1: Tool-using language models with reinforced reasoning},
  author={Zhang, Shaokun and Dong, Yi and Zhang, Jieyu and Kautz, Jan and Catanzaro, Bryan and Tao, Andrew and Wu, Qingyun and Yu, Zhiding and Liu, Guilin},
  journal={arXiv preprint arXiv:2505.00024},
  year={2025}
}

@article{wang2025ragen,
  title={Ragen: Understanding self-evolution in llm agents via multi-turn reinforcement learning},
  author={Wang, Zihan and Wang, Kangrui and Wang, Qineng and Zhang, Pingyue and Li, Linjie and Yang, Zhengyuan and Jin, Xing and Yu, Kefan and Nguyen, Minh Nhat and Liu, Licheng and others},
  journal={arXiv preprint arXiv:2504.20073},
  year={2025}
}

@article{lu2025arpo,
  title={ARPO: End-to-End Policy Optimization for GUI Agents with Experience Replay},
  author={Lu, Fanbin and Zhong, Zhisheng and Liu, Shu and Fu, Chi-Wing and Jia, Jiaya},
  journal={arXiv preprint arXiv:2505.16282},
  year={2025}
}

@article{yan2025memory,
  title={Memory-R1: Enhancing Large Language Model Agents to Manage and Utilize Memories via Reinforcement Learning},
  author={Yan, Sikuan and Yang, Xiufeng and Huang, Zuchao and Nie, Ercong and Ding, Zifeng and Li, Zonggen and Ma, Xiaowen and Sch{\"u}tze, Hinrich and Tresp, Volker and Ma, Yunpu},
  journal={arXiv preprint arXiv:2508.19828},
  year={2025}
}

@article{yang2025qwen3,
  title={Qwen3 technical report},
  author={Yang, An and Li, Anfeng and Yang, Baosong and Zhang, Beichen and Hui, Binyuan and Zheng, Bo and Yu, Bowen and Gao, Chang and Huang, Chengen and Lv, Chenxu and others},
  journal={arXiv preprint arXiv:2505.09388},
  year={2025}
}

@article{li2022alphacode,
  title   = {Competition-level code generation with AlphaCode},
  author  = {Li, Yujia and Choi, David and Chung, Junyoung and Kushman, Nate and Schrittwieser, Julian and Leblond, R{\'e}mi and Eccles, Tom and Keeling, James and Gimeno, Felix and Dal Lago, Agustin and Hubert, Thomas and Choy, Peter and de Masson d’Autume, Cyprien and Babuschkin, Igor and Chen, Xinyun and Huang, Po-Sen and Welbl, Johannes and Gowal, Sven and Cherepanov, Alexey and Molloy, James and Mankowitz, Daniel J. and Robson, Esme Sutherland and Kohli, Pushmeet and de Freitas, Nando and Kavukcuoglu, Koray and Vinyals, Oriol},
  journal = {Science},
  volume  = {378},
  number  = {6624},
  pages   = {1092--1097},
  year    = {2022},
  doi     = {10.1126/science.abq1158}
}

@misc{austin2021program,
  title         = {Program Synthesis with Large Language Models},
  author        = {Austin, Jacob and Odena, Augustus and Nye, Maxwell and Bosma, Maarten and Michalewski, Henryk and Dohan, David and Jiang, Ellen and Cai, Carrie and Terry, Michael and Le, Quoc and Sutton, Charles},
  year          = {2021},
  eprint        = {2108.07732},
  archivePrefix = {arXiv},
  primaryClass  = {cs.PL},
  url           = {https://arxiv.org/abs/2108.07732}
}

@inproceedings{yang-etal-2018-hotpotqa,
  title     = "{H}otpot{QA}: A Dataset for Diverse, Explainable Multi-hop Question Answering",
  author    = "Yang, Zhilin  and Qi, Peng  and Zhang, Saizheng  and Bengio, Yoshua  and Cohen, William  and Salakhutdinov, Ruslan  and Manning, Christopher D.",
  booktitle = "Proceedings of the 2018 Conference on Empirical Methods in Natural Language Processing",
  year      = "2018",
  address   = "Brussels, Belgium",
  publisher = "Association for Computational Linguistics",
  url       = "https://aclanthology.org/D18-1259/",
  doi       = "10.18653/v1/D18-1259",
  pages     = "2369--2380"
}

@article{kwiatkowski-etal-2019-natural,
  title     = "Natural Questions: A Benchmark for Question Answering Research",
  author    = "Kwiatkowski, Tom  and Palomaki, Jennimaria  and Redfield, Olivia  and Collins, Michael  and Parikh, Ankur  and Alberti, Chris  and Epstein, Danielle  and Polosukhin, Illia  and Devlin, Jacob  and Lee, Kenton  and Toutanova, Kristina  and Jones, Llion  and Kelcey, Matthew  and Chang, Ming-Wei  and Dai, Andrew M.  and Uszkoreit, Jakob  and Le, Quoc  and Petrov, Slav",
  journal   = "Transactions of the Association for Computational Linguistics",
  volume    = "7",
  year      = "2019",
  address   = "Cambridge, MA",
  publisher = "MIT Press",
  url       = "https://aclanthology.org/Q19-1026/",
  doi       = "10.1162/tacl_a_00276",
  pages     = "452--466"
}

@misc{shi2025deepresearchsystematicsurvey,
      title={Deep Research: A Systematic Survey}, 
      author={Zhengliang Shi and Yiqun Chen and Haitao Li and Weiwei Sun and Shiyu Ni and Yougang Lyu and Run-Ze Fan and Bowen Jin and Yixuan Weng and Minjun Zhu and Qiujie Xie and Xinyu Guo and Qu Yang and Jiayi Wu and Jujia Zhao and Xiaqiang Tang and Xinbei Ma and Cunxiang Wang and Jiaxin Mao and Qingyao Ai and Jen-Tse Huang and Wenxuan Wang and Yue Zhang and Yiming Yang and Zhaopeng Tu and Zhaochun Ren},
      year={2025},
      eprint={2512.02038},
      archivePrefix={arXiv},
      primaryClass={cs.CL},
      url={https://arxiv.org/abs/2512.02038}, 
}

\appendix
\section*{Appendix}
\section{Dataset Details}
\label{app:datasets}
Table~\ref{tab:datasets} summarizes the datasets used in our experiments and provides the corresponding source links.


\newcommand{\hfds}[1]{\href{https://huggingface.co/datasets/#1}{\mbox{\texttt{#1}}}}

\begin{table*}[t]
\centering
\footnotesize 
\setlength{\tabcolsep}{5pt} 
\renewcommand{\arraystretch}{1.06}
\caption{Datasets used in our experiments.}
\begin{tabularx}{\textwidth}{@{}l l l@{}}
\toprule
\textbf{Domain} & \textbf{Dataset} & \textbf{Link} \\
\midrule
Training (MATH eval) & DAPO-Math-17k-Processed & \hfds{open-r1/DAPO-Math-17k-Processed} \\
\midrule
Math eval & GSM8K & \hfds{openai/gsm8k} \\
Math eval & MATH500 (subset of MATH) & \hfds{HuggingFaceH4/MATH-500} \\
Math eval & AIME24 & \hfds{HuggingFaceH4/aime\_2024} \\
Math eval & AIME25 & \hfds{math-ai/aime25} \\
\midrule
Code analysis & CodeContests & \hfds{dshihk/code-contests} \\
Code analysis & MBPP & \hfds{google-research-datasets/mbpp} \\
\midrule
QA analysis & HotpotQA & \hfds{hotpotqa/hotpot\_qa} \\
QA analysis & Natural Questions (NQ) & \hfds{google-research-datasets/natural\_questions} \\
\bottomrule
\end{tabularx}
\label{tab:datasets}
\end{table*}

\section{Implementation Details}
\label{app:detail}
\subsection{Detailed SEAM Implementations}
\subsubsection{Hyperparameters}
Table \ref{tab:grpo_hparams} and Table \ref{tab:sft_hparams} summarize the key hyperparameters used for GRPO training and deployment-time periodic SFT. Notably, for training efficiency, we decode the executor with zero temperature ($T=0$) during training; accordingly, we set the number of executor rollouts per experience entry to $M=1$.

\begin{table}[t!]
\centering
\small
\caption{Key hyperparameters for GRPO training of SEAM.}
\label{tab:grpo_hparams}
\begin{tabular}{ll}
\toprule
\textbf{Hyperparameter} & \textbf{Value} \\
\midrule
Train batch size & $128$ \\
Max prompt / response length for SEAM& $2048 / 4096$ \\
Max prompt / response length for Executor& $5120 / 8192$ \\
Experience candidates per instance & $8$\\
Executor solutions per instance & $1$\\
Executor temperature & $0.0$ \\
Learning rate  & $1\times 10^{-6}$ \\
KL regularization coef & $0.001$\\
Training epochs & $10$ \\
\bottomrule
\end{tabular}
\label{tab:grpo_hparams}
\end{table}

\begin{table}[t!]
\centering
\small
\caption{Key hyperparameters for deployment-time periodic SFT of SEAM.}
\label{tab:sft_hparams}
\begin{tabular}{ll}
\toprule
\textbf{Hyperparameter} & \textbf{Value} \\
\midrule
Train steps & $500$  \\
Micro-batch size & $8$ \\
Learning rate & $1\times 10^{-5}$ \\
Max prompt / response length & $2048 / 4096$ \\
\bottomrule
\end{tabular}
\label{tab:sft_hparams}
\end{table}

\subsubsection{Prompts}
\label{app:prompt}
In Figure \ref{fig:input_pem} and Figure \ref{fig:executor_prompts}, we present the input prompts for SEAM and the executor.
\subsection{Detailed Baseline Implementations}
\label{app:details}

\subsubsection{General protocol and fairness controls}
\paragraph{Shared frozen executor.}
All baselines use the same frozen executor $\mathcal{E}_{\phi}$ as the final answer generator
(\textsc{Qwen3-4B} or \textsc{Deepseek-R1-distilled-LLaMA-3.1-8B}, matching the main experiments),
so performance differences come from the memory mechanism rather than the executor.

\paragraph{Memory budget and decoding.}
For all memory-augmented baselines, we cap the total memory text concatenated into the executor prompt to 4k tokens.
If a method produces longer memory (e.g., iterative curation), we truncate by keeping higher-priority sections first
(e.g., reusable strategies / formulas) and then cutting from the tail.
All baseline submodules (retrieval, extraction/curation, and final answering) use greedy decoding with temperature $=0.0$.

\paragraph{Online evaluation and run averaging.}
\textsc{Dynamic-Cheatsheet}, \textsc{Mem-0}, and \textsc{Memento} can update memory online and may be order-sensitive.
Thus, for each dataset we sample a fresh random permutation of the evaluation set per run, reset to the same prefilled bank using the same training set of SEAM,
process instances sequentially, and report the mean over 10 independent runs.

\subsubsection{\textsc{Mem-0} (graph-based long-term memory)}
\paragraph{Adaptation to math reasoning.}
\textsc{Mem-0} is originally designed for incremental conversational memory.
We adapt it by treating each processed training/evaluation instance as a single interaction unit:
(user message = problem statement; assistant message = model solution).
The external memory stores structured ``facts'' that can help future problems, such as reusable lemmas,
problem templates, common transformations, and frequent error patterns.

\paragraph{Two-stage pipeline: extraction and update.}
We follow the standard \textsc{Mem-0} pipeline consisting of:
(i) \emph{extraction}, which extracts a set of salient candidate memories from the latest interaction
given (a) a global summary of the history and (b) a recency window of recent interactions;
(ii) \emph{update}, which retrieves semantically similar existing memories and decides one of four actions:
\textsc{ADD}, \textsc{UPDATE}, \textsc{DELETE}, or \textsc{NOOP} for each candidate memory, to maintain consistency and avoid redundancy.

\paragraph{Hyperparameters.}
We use a recency window $m=10$ interactions and retrieve $s=10$ similar memories during the update phase.
All similarity search uses dense embeddings with cosine similarity.
After the memory is updated, we retrieve the top-$r$ memories most relevant to the current problem (default $r=10$),
serialize them into text, and prepend them to the executor prompt as additional context
(while respecting the 4k-token memory budget).

\subsubsection{\textsc{Dynamic-Cheatsheet} (DC-RS)}
\paragraph{Method overview.}
We adopt the DC-RS (retrieve-and-synthesize) variant, which maintains a persistent text cheatsheet (memory)
and updates it \emph{before} solving each new problem.
At step $i$, DC-RS performs:
(1) retrieve top-$k$ most similar past inputs and their model-generated outputs,
(2) update the cheatsheet by a \emph{curator} conditioned on the previous cheatsheet and retrieved pairs,
and (3) solve the current problem by a \emph{generator} conditioned on the updated cheatsheet.

\paragraph{Retrieval and prompts.}
We set $k=3$ for retrieval.
The retriever ranks historical inputs by embedding cosine similarity and returns the top-$k$ input--output pairs.
We follow DC's prompt separation between a generator (problem solver) and a curator (memory editor):
the generator prompt contains a \texttt{[[CHEATSHEET]]} placeholder and a \texttt{[[QUESTION]]} placeholder;
the curator prompt takes the previous cheatsheet and retrieved notes and outputs an updated cheatsheet.
We keep the cheatsheet as a structured document (e.g., reusable strategies / heuristics) and enforce the global 4k-token memory budget.

\paragraph{Online update without labels.}
Consistent with DC, the curator does not use ground-truth answers.
After each problem is answered, we append the (problem, model output) pair to the retriever store;
the next step may retrieve it to refine the cheatsheet.

\subsubsection{\textsc{Memento} (case-based reasoning with a case bank)}
\paragraph{Adaptation to single-step math problems.}
\textsc{Memento} is proposed as a planner--executor agent with a growing case bank.
For math benchmarks without tool interaction, we instantiate the core case-based reasoning (CBR) component
in a single-step setting:
each problem is treated as a state $s$, the model solution as an action $a$, and the scalar reward $r$
is computed from answer correctness.

\paragraph{Case bank (Write/Read).}
We use the \emph{non-parametric} case memory variant.
\textbf{Write} appends each completed case $(s,a,r)$ to the case bank.
\textbf{Read} retrieves top-$K$ nearest cases by cosine similarity between a frozen text-encoder representation of the current state
and stored past states.

\paragraph{Hyperparameters and prompting.}
We set $K=3$ to control context length and match the retrieval scale of other baselines.
Retrieved cases are formatted as short exemplars (problem statement + key steps + final answer, optionally with the reward tag),
then concatenated to the executor prompt as reference experiences.
We again enforce the 4k-token memory budget for the concatenated exemplars.

\paragraph{Why we use non-parametric CBR.}
While \textsc{Memento} also supports a parametric retrieval policy (online-updated Q-function),
we use the non-parametric variant to avoid introducing an additional trainable module and to keep the comparison focused on retrieval-augmented reuse.

\begin{figure*}[t]
\centering
\begin{tcolorbox}[
  title=Input prompt for SEAM,
  colback=boxbg,
  colframe=black,
  coltitle=white,
  colbacktitle=titlebg,
  fonttitle=\bfseries,
  sharp corners=south,
  boxrule=0.8pt,
  width=\textwidth
]
\small
{\ttfamily
You are an Experience Generator that produces structured guidance for a downstream large-language executor.\\
Here is the information you receive:\\
\\
<problem>\\
\{problem\}\\
</problem>\\
\\
Using this information, create one structured experience entry with three parts:\\
1. <analysis> -- a brief, executor-aware diagnosis of what the problem is asking and what is likely to be tricky (e.g., key constraints, typical failure modes, and the most suitable strategy family). Keep it short.\\
2. <experience> -- a bullet list of distilled tips designed for the downstream executor that will help it solve the current problem <problem> \{problem\} </problem> and similar tasks in the future. Draw on the executor's typical strengths and weaknesses to give algorithms, heuristics, and pitfalls that will most help it on future attempts.\\
3. <example> -- 3--8 concise, high-level steps you judge to be the most effective solution path for the executor on the problem <problem> \{problem\} </problem>.\\
---\\
\\
\#\# Output Format\\
\\
<analysis>\\
\ldots brief diagnosis and strategy\\
</analysis>\\
\\
<experience>\\
\textbullet\ \ldots first distilled insight\\
\textbullet\ \ldots second distilled insight\\
\textbullet\ \ldots etc.\\
</experience>\\
\\
<example>\\
1.\ Step 1 -- \ldots\\
2.\ Step 2 -- \ldots\\
3.\ \ldots\\
</example>\\
}
\end{tcolorbox}

\caption{Input prompt for SEAM.}
\label{fig:input_pem}
\end{figure*}


\begin{figure*}[t]
\centering

\begin{subfigure}[t]{\textwidth}
\centering
\begin{tcolorbox}[
  title=Input prompt for the executor Qwen3,
  colback=boxbg,
  colframe=black,
  coltitle=white,
  colbacktitle=titlebg,
  fonttitle=\bfseries,
  sharp corners=south,
  boxrule=0.8pt,
  width=\textwidth
]
\small\ttfamily
The above is a Q\&A dialogue between a user and a problem-solving guidance model.\\
Treat the output of the guidance model as advisory context to solve the problem: prefer using its techniques when they fit, but you may use alternative correct methods if they are more efficient or clearer. If you diverge from the advisory context, briefly explain why. Be concise and accurate.\\
Present your reasoning and answer in the following format:\\
\verb|<think> Content of Thinking</think><answer>[Final result only]</answer>|
\end{tcolorbox}
\caption{Qwen3 executor.}
\label{fig:prompt_qwen3}
\end{subfigure}

\vspace{0.6em}

\begin{subfigure}[t]{\textwidth}
\centering
\begin{tcolorbox}[
  title=Input prompt for the executor Deepseek-R1-distilled,
  colback=boxbg,
  colframe=black,
  coltitle=white,
  colbacktitle=titlebg,
  fonttitle=\bfseries,
  sharp corners=south,
  boxrule=0.8pt,
  width=\textwidth
]
\small\ttfamily
The above is a Q\&A dialogue between a user and a problem-solving guidance model.\\
Treat the output of the guidance model as advisory context to solve the math problem: prefer using its techniques when they fit, but you may use alternative correct methods if they are more efficient or clearer.\\
If you diverge from the advisory context, briefly explain why. Be concise and accurate.\\
Please reason step by step, and put your final answer within \verb|\boxed{}|.
\end{tcolorbox}
\caption{Deepseek-R1-distilled executor.}
\label{fig:prompt_deepseek}
\end{subfigure}

\caption{Input prompts for different executors.}
\label{fig:executor_prompts}
\end{figure*}

\section{Case Study}\label{app:case}
To examine the behavior of our method on real task instances, we selected a geometry problem from Math500 as a case study. The problem is: in triangle \(ABC\), with \(AB=17\), \(AC=8\), and \(BC=15\), let \(D\) be the foot of the altitude from \(C\) to \(AB\). Find the area of \(\triangle ACD\). The correct answer is \(\tfrac{3840}{289}\).

In this case, the \texttt{analysis} section of SEAM provided a concise and correct problem analysis: it first verified that \(\triangle ABC\) is right-angled at \(C\) by the Pythagorean theorem, and then suggested placing the triangle in a coordinate system (e.g., \(C=(0,0),\,A=(0,8),\,B=(15,0)\)), finding the coordinates of \(D\), and finally computing the area of \(\triangle ACD\) using the shoelace formula (or equivalently, base–height/2).  
The \texttt{experience} section listed multiple possible solution strategies (similar triangles, area decomposition, coordinate geometry, etc.) and explicitly recommended the coordinate method as the most direct, efficient, and reliable approach for this problem.  
The \texttt{example} section further provided the concrete steps of the coordinate method (deriving the equation of line \(AB\), the perpendicular line through \(C\), their intersection point \(D\), and then applying the area formula), yielding a reusable “action checklist.”

The downstream executor successfully solved the problem and obtained the correct result \(\tfrac{3840}{289}\). Notably, the executor was not additionally fine-tuned, and thus exhibited a degree of autonomy: it initially attempted two alternative approaches (e.g., area ratios/similar triangles and direct algebraic decomposition), but after cross-checking intermediate results, it eventually followed SEAM’s suggestion, switched to the simpler coordinate method, and completed the exact computation. This illustrates two points: (i) SEAM outputs provide a clear strategic prior that guides the executor toward convergence when multiple strategies compete, and (ii) even without specialized training, the executor may perform exploratory reasoning, but under the structured guidance of SEAM, it ultimately converges to the most efficient and verifiable solution path.

This case study demonstrates that SEAM’s structured experience and task decomposition not only improve efficiency (by reducing unproductive exploration) but also enhance interpretability and reusability of reasoning steps. The executor’s limited autonomous exploration does not hinder final correctness; rather, in combination with SEAM, it forms a cooperative paradigm of “prior guidance + verification-based convergence.” This division of labor allows SEAM to focus on \emph{experiential strategy generation}, while the executor specializes in \emph{step execution and verification}. Without additional fine-tuning, the system can still produce stable and correct solutions on benchmark problems.

\begin{tcolorbox}[breakable, colback=gray!3, colframe=gray!40, title={Case Study}]
\small
\textbf{Original Problem}\\
{\ttfamily
Given triangle \(ABC\) with \(AB=17\), \(AC=8\), \(BC=15\). Let \(D\) be the foot of the altitude from \(C\) to \(AB\).
Find the area of triangle \(ACD\).
}
\medskip\hrule\medskip

\textbf{SEAM Output}\\
{\ttfamily
<analysis>\\
The problem asks for the area of triangle ACD. In triangle ABC, the side lengths are AB = 17, AC = 8, and BC = 15, and D is the foot of the altitude from C to AB.\\

First, I need to determine the type of triangle ABC. Check the Pythagorean relation:
\(8^2 + 15^2 = 64 + 225 = 289 = 17^2\), so triangle ABC is right-angled at C.\\

This lets me place the triangle on a coordinate plane for convenience. Set \(C\) at the origin \((0,0)\), \(A\) at \((0,8)\), and \(B\) at \((15,0)\). Then draw the perpendicular from \(C\) to \(AB\) and find the coordinates of the foot \(D\).\\

With the coordinates of \(A\), \(C\), and \(D\) in hand, I can compute the area of triangle \(ACD\) using the shoelace formula or the ``base \(\times\) height / 2'' formula. This is more direct and efficient than using trigonometry or similar triangles for this problem.\\
</analysis>
}

\medskip

{\ttfamily
<experience>\\
\textbullet\ Identify the triangle type: use the Pythagorean theorem to check if it is a right triangle. Given \(AB=17\), \(AC=8\), and \(BC=15\), verify
\(8^2+15^2=64+225=289=17^2\); therefore, triangle \(ABC\) is right-angled at \(C\).\\
\textbullet\ Use coordinate geometry: place \(C\) at \((0,0)\), \(A\) at \((0,8)\), and \(B\) at \((15,0)\), then find the coordinates of \(D\), the foot of the altitude from \(C\) to \(AB\).\\
\textbullet\ Use the triangle area formula: since \(ABC\) is a right triangle, its area is
\(\tfrac{1}{2}\times AC \times BC=\tfrac{1}{2}\times 8 \times 15=60\).\\
\textbullet\ Find the area of triangle \(ACD\) using coordinates: with \(A\), \(C\), and \(D\), apply the shoelace formula or the ``base \(\times\) height / 2'' formula.\\
\textbullet\ Similar triangles or trigonometry could also be used, but in this case, coordinate geometry is more efficient.\\
</experience>
}

\medskip

{\ttfamily
<example>\\
1.\ Step 1 --- Use the Pythagorean theorem to determine whether triangle \(ABC\) is right-angled.\\
2.\ Step 2 --- Place the triangle on a coordinate plane with \(C\) at \((0,0)\), \(A\) at \((0,8)\), and \(B\) at \((15,0)\).\\
3.\ Step 3 --- Find the coordinates of \(D\), the foot of the altitude from \(C\) to \(AB\).\\
4.\ Step 4 --- Using the coordinates of \(A\), \(C\), and \(D\), compute the area of triangle \(ACD\) with the shoelace formula.\\
</example>
}

\medskip\hrule\medskip

\textbf{Executor Output}\\
{\ttfamily
<think>\\
We are asked to find the area of triangle \(ACD\) inside triangle \(ABC\), where \(AB=17\), \(AC=8\), \(BC=15\), and \(D\) is the foot of the altitude from \(C\) to \(AB\).

First, the user mentioned that the triangle is a right triangle. Let me confirm that. Using the Pythagorean theorem: \(AC^2 + BC^2\) should equal \(AB^2\). \(8^2\) is 64, \(15^2\) is 225, adding them gives 289, which is \(17^2\). So yes, triangle ABC is right-angled at C. That simplifies things because the area of ABC is \((AC \times BC)/2 = (8 \times 15)/2 = 60\).\\

Now, the altitude from C to AB. In a right triangle, the altitude from the right angle to the hypotenuse has a known length. The formula for the altitude (h) is (product of the legs)/hypotenuse. So \(h = (8 \times 15)/17 = 120/17\). But wait, that's the length of the altitude from C to AB, which is CD.\\

But the question asks for the area of triangle ACD. Since D is the foot of the altitude, CD is the height of triangle ACD with base AD. So I need to find AD.\\

In a right triangle, the length of the segments into which the altitude divides the hypotenuse can be found using similar triangles. The formula for AD is \((AC^2)/AB\). So \(AD = (8^2)/17 = 64/17\).\\

Then the area of ACD is \((AD \times CD)/2\). Plugging in the values: \((64/17 \times 120/17)/2 = (7680/289)/2 = 3840/289\). Let me check that calculation again.\\

Alternatively, since the area of ABC is 60, and D divides AB into AD and DB. The area of ACD would be proportional to \(AD/AB\) times the area of ABC. So \(60 \times (AD/AB) = 60 \times (64/17)/17 = 60 \times 64/(17^2) = 3840/289\). That's the same result.\\

So the area of triangle ACD is \(3840/289\). Let me compute that as a decimal to check if it makes sense. 289 goes into 3840 about 13.3 times. But since the problem likely expects an exact fraction, \(3840/289\) is the answer. Wait, but maybe there's a simpler way.\\
It is suggested that the coordinate method is simpler, so let me give it a try. Use coordinates: set \(C=(0,0)\), \(A=(0,8)\), \(B=(15,0)\).

Line \(AB\) passes through \((0,8)\) and \((15,0)\), so its slope is \(-\frac{8}{15}\) and its equation is
\[
y=-\frac{8}{15}x+8.
\]
The altitude from \(C\) is perpendicular to \(AB\), hence has slope \(\frac{15}{8}\) and equation
\[
y=\frac{15}{8}x.
\]
Intersecting these gives
\[
\begin{aligned}
\frac{15}{8}x &= -\frac{8}{15}x + 8 
\;\Longrightarrow\; 289x=960,\\
x &= \frac{960}{289}, 
\qquad
y=\frac{15}{8}x=\frac{1800}{289}.
\end{aligned}
\]

Thus
\[
D=\left(\frac{960}{289},\,\frac{1800}{289}\right).
\]

Compute the area of \(\triangle ACD\) using the determinant (shoelace) formula with
\(A(0,8)\), \(C(0,0)\), \(D\!\left(\frac{960}{289},\frac{1800}{289}\right)\):
\[
\begin{aligned}
&[ACD]\\[-0.2em]
&= \frac12\left|x_A(y_C-y_D)+x_C(y_D-y_A)+x_D(y_A-y_C)\right| \\
&= \frac12\cdot\frac{960}{289}\cdot 8
= \frac{3840}{289}.
\end{aligned}
\]


Therefore, the area is \(\boxed{\frac{3840}{289}}\).
</think>

<answer> \( \dfrac{3840}{289} \)</answer>
}

\end{tcolorbox}

\end{document}